\documentclass[letterpaper, 10 pt, conference]{ieeeconf}  

\IEEEoverridecommandlockouts                              

\usepackage{amsmath,amssymb,xspace}
\usepackage{graphicx}
\usepackage{float}
\usepackage{algorithm}
\usepackage{algorithmic}
\usepackage{verbatim}
\usepackage{dsfont}
\usepackage{enumerate}
\usepackage{varwidth}
\usepackage{multirow}
\usepackage{subfigure}

\DeclareMathOperator{\ad}{ad}
\DeclareMathOperator{\Ad}{Ad}
\DeclareMathOperator{\ID}{ID}
\DeclareMathOperator{\FD}{FD}
\newcommand\SE[1][3]{\ensuremath{\mathrm{SE}(#1)}\xspace}
\newcommand\se[1][3]{\ensuremath{\mathfrak{se}(#1)}\xspace}
\newcommand\GL[1]{\ensuremath{\mathrm{GL}_{\mathds R}(#1)}\xspace}

\newcommand\algorithmicparcmpt{\textbf{Parallel Compute:}}
\newcommand\PARCMPT{\item[\algorithmicparcmpt]}

\newcommand\algorithmicfwscan{\textbf{Forward Scan:}}
\newcommand\FWSCAN{\item[\algorithmicfwscan]}

\newcommand\algorithmicbwscan{\textbf{Backward Scan:}}
\newcommand\BWSCAN{\item[\algorithmicbwscan]}

\newcommand\algorithmichetercomp{\textbf{Asynchronous Compute:}}
\newcommand\HETER{\item[\algorithmichetercomp]}

\newcommand\algorithmicstreampar{\textbf{Stream Parallel Compute:}}
\newcommand\STREAM{\item[\algorithmicstreampar]}

\title{\LARGE \bf
Parallel Dynamics Computation using Prefix Sum Operations
}

\author{Yajue Yang \and Yuanqing Wu \and Jia Pan
\thanks{Yajue Yang and Jia Pan are with the Department of Mechanical and Biomedical Engineering, the City University of Hong Kong; Yuanqing Wu is with the University of Bologna, Italy
        }%
}

\begin{document}

\maketitle
\thispagestyle{empty}
\pagestyle{empty}

\begin{abstract}
	We propose a new parallel framework for fast computation 
	of inverse and forward dynamics of articulated robots 
	based on prefix sums (scans).	
	We re-investigate the well-known recursive Newton-Euler 
	formulation of robot dynamics and show that the forward-backward 
	propagation process for robot inverse dynamics is equivalent 
	to two scan operations on certain semigroups.
	We show that the state-of-the-art forward dynamics algorithms may almost completely be 	cast into a sequence of scan operations, with unscannable parts clearly identified. This suggests a serial-parallel hybrid approach for systems with a moderate number of links.
We implement our scan based algorithms on Nvidia CUDA platform with performance compared with multithreading CPU-based recursive algorithms; a significant level of acceleration is demonstrated.

\end{abstract}

\begin{table*}
\begin{tabular}{p{.08\textwidth}p{.37\textwidth}p{.04\textwidth}p{.08\textwidth}p{.37\textwidth}}
\SE & Euclidean group of $\mathds R^3$   & & \se & Lie algebra of \SE\\
$f_{i,i+1}$ & \mbox{coordinate transformation} from link frame $i+1$ to $i$ & & $M_i$ & initial coordinate transformation of $f_{i-1,i}$ \\
$\Ad_{(\cdot)}$ & Adjoint transformation & & $\ad_{(\cdot)}$ & adjoint map\\
$S_i$ & $i^{\text{th}}$ joint twist & & $S$ & $\text{diag}(S_1,\dots,S_n)$\\
$q_i$ & $i^{\text{th}}$ joint variable & & $q$ & $(q_1,\dots,q_n)^T$\\
$V_i$ & spatial velocity of $i^{\text{th}}$ link & & $V$ & $\text{column}(V_1,\dots,V_n)$ \\
$\dot V_i$ & spatial acceleration of $i^{\text{th}}$ link & & $\dot V$ & $\text{column}(\dot V_1,\dots,\dot V_n)$ \\
$F_i$ & reaction force of $i^{\text{th}}$ joint & & $F$ & $\text{column}(F_1,\dots,F_n)$ \\
$\tau_i$ & $i^{\text{th}}$ joint torque & & $\tau$ & $(\tau_1,\dots,\tau_n)^T$ \\
$J_i$ & $i^{\text{th}}$ link inertia & & $J$ & $\text{diag}(J_1,\dots,J_n)$\\
$\hat J_i$ & $i^{\text{th}}$ articulated body inertia & & $\hat J$ & $\text{diag}(\hat J_1,\dots,\hat J_n)$\\
$M(q)$ & joint space inertia at $q$ & & $\mathds R^{n\times n}$ & $n^{\text{th}}$-order matrix space \\
$\GL{n}$ & \mbox{~~~$n^{\text{th}}$-order general linear group} &&$\oplus$& binary associative operator\\
$[x_i]$ & array $[x_1,\dots,x_n]$ &&&
\end{tabular}
\caption{nomenclature}
\end{table*}

\section{Introduction}
\label{sec:intro}

Recent developments in humanoid robot online motion planning/replanning ~\cite{yamane2003dynamics,lee2005newton,chitchian2012particle,lengagne2013generation,Chretien16} and model-based control optimization and learning~\cite{Todorov:2012:Mujoco,Erez:2012:TOD} have raised new computational challenges on classical articulated robot dynamics algorithms: hundreds to thousands of groups of inverse/forward dynamics and their derivatives under different states and control inputs must be evaluated within several milliseconds. Fortunately, such computation task is inherently akin to parallel computation on multiple levels, and therefore may take substantially less time than a sequential implementation. At the outset, data-independent parallelism from different states and inputs suggests a parallel computation on the group level. Within individual groups, the data-dependent parallelism in inverse and forward dynamics is well known to the robotics community, thanks to decades of extensive research on adapting recursive sequential algorithms ~\cite{featherstone1983calculation,rodriguez1991spatial,park1995lie} to parallel algorithms for multi-processor systems ~\cite{fijany1995parallel,featherstone1999divide1,featherstone1999divide2,anderson2000highly,yamane2007automatic}. Such algorithms are mainly intended for complex simulations with a large number (e.g., $\geq 1000$) of links that are unusual for robotic applications. 

A realistic scenario for dynamics computation in articulated robotics may involve simultaneously solving a large number of groups ($>1000$) of inverse/forward dynamics problem for a robot with a moderate number ($10\sim 100$) of links~\cite{lengagne2013generation,Chretien16,Todorov:2012:Mujoco,Erez:2012:TOD,laflin2016enhancing,fijany2013new}. It is arguably more suitable for a hybrid CPU/GPU implementation in light of the recent popularization of GPGPU technology~\cite{Nick10}. Several researchers~\cite{Chretien16,Zhang98} have identified part of the inverse dynamics as a prefix sum operation, also known as scan~\cite{blelloch1990prefix,harris2007parallel}, which is a useful and easy to implement building block for many parallel algorithms and is well supported by GPU platforms like NVidia CUDA~\cite{bell2011thrust}. These early work does not exploit the Lie group structure within the inverse dynamics, and thus needs to compute a large set of temporary parameters in parallel to prepare the data before applying a parallel prefix sum. In fact, articulated robot kinematics and dynamics can be considered a spatial propagation problem~\cite{rodriguez1991spatial,park1995lie,rodriguez1987kalman,featherstone2014rigid}, which naturally leads to linear recursions equivalent to scan operations~\cite{blelloch1990prefix}. This implies that the inverse/forward dynamics problem may be mostly (if not entirely) modeled as a sequence of scan operations, thereby applying state-of-the-art parallel computing technology to the classic dynamics computation problem without reinventing any wheels~\cite{negrut2014parallel}.

In this paper, we will show that the parallel scan perspective on robot dynamics computation can indeed be carried to a further extent: apart from the initialization and some intermediate variables that may be computed in parallel, both the inverse dynamics and the forward dynamics problem may be reformulated into a sequence of scan operations. The paper is organized as follows. In Section~\ref{sec:dyn-recap}, we give a brief review of the recursive Newton-Euler formulation for robot dynamics, following the Lie group approach formally stated in~\cite{park1995lie}. Our main contribution is presented in Section~\ref{sec:dyn-scan} and~\ref{sec:gpu-impl}, where we formulate the recursive inverse and forward dynamics problem into a sequence of scan operations. We show that the scan operator can be considered as the binary operation of certain semigroups. Detailed algorithms of implementations are then described.
In Section~\ref{sec:experiment}, we present some experiment results of a hybrid CPU/GPU implementation of our parallel scan based algorithm and make some reasonable comparison with (multithreading) CPU based implementations. In particular, our method achieves a maximal 500x speedup on inverse dynamics computation, and a maximal 100x speedup on forward dynamics. Without loss of generality, we shall only consider single open chain robot in this paper.

\section{Lie Group Formulation of Robot Dynamics}
\label{sec:dyn-recap}
\subsection{Inverse dynamics}
\label{sec:recap:ID}

A comprehensive summary of state-of-the-art robot inverse and forward dynamics algorithms is given by~\cite{featherstone2014rigid}. The inverse dynamics problem of deriving necessary joint torques for generating a specified robot motion is well-known to be a forward-backward two-phase propagation process ~\cite{park1995lie,rodriguez1987kalman,murray1994mathematical}. In the forward propagation phase, the motion of each joint (with twist axis $S_i$ and joint variables $q_i$, $\dot{q}_i$ and $\ddot{q}_i$) propagates from the base link toward the end-effector of a $n$-link robot, resulting in linear recursion of link velocities and accelerations for $i=1,\dots,n$:
\begin{eqnarray}\left\{ 
	\begin{aligned}
		f_{i-1,i}&=M_ie^{S_iq_i}\\
		V_i&=\Ad_{f_{i-1,i}^{-1}}( V_{i-1})+S_i\dot q_i\\
		\dot V_i&=S_i\ddot q_i+\Ad_{f_{i-1,i}^{-1}}( \dot V_{i-1} )-\ad_{S_i\dot q_i}\Ad_{f_{i-1,i}^{-1}}(V_{i-1}).
	\end{aligned}\right.
	\label{eq:id-forward}
\end{eqnarray}
Here, we follow closely the notation of ~\cite{park1995lie,murray1994mathematical,ploen1999coordinate}. In the back propagation phase, the reaction force applied by each joint to the succeeding link is successively computed using Newton-Euler equation, which leads to a second set of linear recursion for joint reaction forces and joint torques for $i=n,\dots,1$:
\begin{equation} 
	\left\{  
		\begin{aligned}
			F_i&=\Ad_{f_{i,i+1}^{-1}}^T(F_{i+1})+J_i\dot V_i-\ad_{V_i}^T(J_iV_i)\\
			\tau_i&=S_i^TF_i.
		\end{aligned}
		\right.	
	\label{eq:id-backward}
\end{equation}
The base link velocity and acceleration $V_0,\dot V_0$ and external force $F_{n+1}$ acting on the end link are provided to initiate the recursion. A sequential recursion of \eqref{eq:id-forward} and \eqref{eq:id-backward} leads to the $O(n)$ (for $n$ links) recursive Newton-Euler inverse dynamics algorithm. Here, emphasis should be made on the coordinate invariant Lie group description of robot kinematics and dynamics ~\cite{murray1994mathematical,ploen1999coordinate}. We will show in Section \ref{sec:dyn-scan} that the scan operands and operators defined by \eqref{eq:id-forward} and \eqref{eq:id-backward} may be considered as the elements and binary operator of certain matrix semigroup that is closely related to \SE (special Euclidean group of $\mathds R^3$). This makes it possible to adapt our parallel framework to using other mathematical representations of \SE, such as dual quaternions ~\cite{dooley1991spatial} or other Clifford algebras ~\cite{selig2010rigid}. For the rest of the paper, we shall denote the ID function by:
\begin{equation}
	\tau=\ID(q,\dot q,\ddot q,V_0,\dot V_0,F_{n+1}).
	\label{}
\end{equation}

\subsection{Forward dynamics}
\label{sec:racap:FD}

The robot forward dynamics (FD) problem refers to the derivation of joint acceleration $\ddot q=(\ddot q_1,\dots,\ddot q_n)^T$ as a function of the states $(q^T,\dot q^T)^T$, applied joint torques $\tau$, base link velocity and acceleration $V_0,\dot V_0$, and external force $F_{n+1}$:
\begin{equation}
	\ddot q=\FD(q,\dot q,\tau,V_0,\dot V_0,F_{n+1})
	\label{eq:fd}
\end{equation}
It may be numerically integrated to compute joint position and velocity trajectories. The joint torques $\tau$ is linearly related to the joint accelerations $\ddot q$ via the \emph{joint space inertia} (JSI) $M(q)$ with a bias term $\tau^{\text{bias}}$ accounting for Coriolis, centrifugal and external forces: 
\begin{equation}\left\{  
	\begin{aligned}
		\tau=&M(q)\ddot q+\tau^{\text{bias}}\\
		\tau^{\text{bias}}:=&\ID(q,\dot q,0,V_0,\dot V_0,F_{n+1})
	\end{aligned}\right.
	\label{eq:tau-bias}
\end{equation}
Since both the joint accelerations $\ddot q_i$'s and joint reaction forces $F_i$'s are unknown, the original FD problem obstructs a propagation formulation and cannot be directly solved by parallel scan operations. However, $\ddot q$ may be derived from directly inverting the JSI,  
\begin{equation}
	\ddot q=M^{-1}(q)\hat\tau:=M^{-1}(q)(\tau-\tau^{\text{bias}})
	\label{eq:JSIinv}
\end{equation}
leading to the $O(n^3)$ \emph{JSI inversion algorithm} (JSIIA) ~\cite[ch. 6]{featherstone2014rigid}. Depending on the methods for computing JSI, JSII may be computationally appealing for moderate number of links ($n\leq 200$). However, it does not take advantage of the structure of JSI inherited from robot chain topology. This was emphasized in a series of papers on analytical factorization and inverse of JSI ~\cite{rodriguez1991spatial,fijany1995parallel,fijany2013new,rodriguez1987kalman,jain1991unified}. The goal of the factorization is to represent the inverse of JSI as the product of block diagonal and block upper/lower triangular matrices, or a sequence of linear recursions that leads to the computationally efficient \emph{propagation} methods~\cite[ch. 7]{featherstone2014rigid}. 

The \emph{articulated-body inertia algorithm} (ABIA) proposed by Featherstone~\cite{featherstone1983calculation} is a well-known propagation method that involves a nonlinear recursion for deriving the equivalent inertias of the sub-system rooted at link $i$ (see ~\cite{ploen1999coordinate} for the notations) for $i=n,\dots,1$:
\begin{equation}
\begin{aligned}
	\hat J_i&=&J_i&+\Ad_{f_{i,i+1}^{-1}}^T\hat J_{i+1}\Ad_{f_{i,i+1}^{-1}} \\
    &&-&\frac{\Ad_{f_{i,i+1}^{-1}}^T\hat J_{i+1}S_{i+1}S_{i+1}^T\hat J_{i+1}\Ad_{f_{i,i+1}^{-1}}}{S_{i+1}^T\hat J_{i+1}S_{i+1}}	
 \end{aligned}
	\label{eq:abi}
\end{equation}
which is essentially a recursive elimination of the unknown joint reaction forces $F_i$'s using virtual work principle. The availability of ABI allows us to derive link accelerations from joint torques $\tau_i$'s with a backward-forward two-phase propagation process~\cite{featherstone2014rigid}, which may be summarized as follows \cite{ploen1999coordinate}:
\begin{equation}
\begin{array}{c}
\text{Backward recursion} \\
(\text{for }i=n,\dots,1) \\
\\
\left\{  
	\begin{aligned}
		\hat z_i&=Y_{i,i+1}\hat z_{i+1}+\mathrm\Pi_{i,i+1}\hat\tau_{i+1}\\
		c_i&=\hat\tau_i-S_i^T\hat z_i\\
		\hat c_i&=\mathrm\Omega_i^{-1}c_i:=(S_i^T\hat J_iS_i)^{-1}c_i
	\end{aligned}\right.\\
    \\
\text{Forward recursion}\\
(\text{for }i=1,\dots,n)\\
\\
\left\{  
	\begin{aligned}
		\lambda_i&=Y_{i-1,i}^T\lambda_{i-1}+S_i\hat c_i\\
		\ddot q_i&=\hat c_i-\mathrm\Pi_{i-1,i}^T\lambda_{i-1}
	\end{aligned}\right.\\
 \end{array}
	\label{eq:recursive-fd}
\end{equation}
where $\displaystyle Y_{i,i+1}:=\Ad_{f_{i,i+1}^{-1}}^T\left( I-\frac{\hat J_{i+1}S_{i+1}S_{i+1}^T}{S_{i+1}^T\hat J_{i+1}S_{i+1}} \right)$ and $\displaystyle \mathrm\Pi_{i,i+1}:=\frac{\Ad_{f_{i,i+1}^{-1}}^T\hat J_{i+1}S_{i+1}}{S_{i+1}^T\hat J_{i+1}S_{i+1}}$.

The connection of ABI to square factorization of JSI is emphasized in~\cite{rodriguez1991spatial}. All recursions involved in ABIA except that of \eqref{eq:abi} are linear and may eventually be formulated as scan operations. However, eq. \eqref{eq:abi} is a well known nonlinear recursion that does not speedup beyond a constant factor by parallel algorithms ~\cite{miklovsko1984complexity,hyafil1977complexity}, and therefore should not conform to a parallel scan operation. The joint reaction forces may also be eliminated by defining appropriate (not necessarily orthogonal) complements for joint constraint force, leading to the \emph{constraint force algorithm} (CFA)~\cite{fijany1995parallel}; the inverse of JSI is directly factorized into a product of block diagonal, block upper/lower bi-diagonal and block tri-diagonal matrices. Block cyclic-reduction algorithms~\cite{heller1976some,sweet1977cyclic} may be applied to solve a block tri-diagonal system efficiently.
\section{Prefix-Sum Re-formulation of Robot Dynamics}
\label{sec:dyn-scan}

\subsection{A brief review of scan operation}

The all-prefix-sums (scan) operation~\cite{blelloch1990prefix} takes a binary associative operator $\oplus$, and an array of $n$ elements
\begin{equation}
	\left[ a_0, a_1, \dots, a_{n-1}\right]	
	\label{}
\end{equation}
and returns the array
\begin{equation}
\begin{aligned}
	\left[ x_0, x_1,\dots, x_{n-1}\right]:= & [ a_0, (a_0 \oplus a_1),\dots, \\
    & (a_0 \oplus a_1 \oplus\cdots\oplus a_{n-1})]
    \end{aligned}
	\label{eq:scan}
\end{equation}
It is obvious that the scan operation implies the following recursion:
\begin{equation}
	x_i=x_{i-1} \oplus a_i
	\label{}
\end{equation}
Generalizations of the standard scan operation and their applicatinos are discussed in~\cite{blelloch1990prefix}. 

\subsection{Synchronous forward scan of link velocities and accelerations}
\vskip-10pt

Our following development is inspired by Zhang \emph{et al.}'s work~\cite{Zhang98} on scanning successive multiplications of link rotation transformations. Its practical application in humanoid motion planning is considered in~\cite{Chretien16}. More generally, we may rewrite the second and third equation of \eqref{eq:id-forward} in a single linear recursion:
\begin{equation}
\begin{aligned}
	\begin{bmatrix}
		\dot V_i\\
		V_i\\
		\hline
		1
	\end{bmatrix}&=\underbrace{\left[
	\begin{array}{cc|c}
		\Ad_{f_{i-1,i}^{-1}} &  -\ad_{S_i\dot q_i}\Ad_{f_{i-1,i}^{-1}} & S_i\ddot q_i\\
				 0   & \Ad_{f_{i-1,i}^{-1}} & S_i\dot q_i\\
				    \hline
				  0  &        0         & 1
	\end{array}
\right]}_{\displaystyle A_i\in\GL{13}}\begin{bmatrix}
		\dot V_{i-1}\\
		V_{i-1}\\
		\hline
		1
	\end{bmatrix}\quad \\ A_0&=\left[\begin{array}{cc|c}
	I & 0 & \dot V_0\\
    0 & I & V_0\\
    \hline
    0 & 0 & 1
\end{array}\right]
\end{aligned}
	\label{eq:scan-id-vel-acc}
\end{equation}
leading to a synchronous scan of both link velocities $V_i$'s and accelerations $\dot V_i$'s. Emphasis should be made on the Lie group structure of the operands: it can be shown that the operands are isomorphic to $(f_{i-1,i}^{-1},S_i\ddot q_i,S_i\dot q_i)\in\SE\times\se^2$, a Lie group with binary operation $\oplus$ defined by:
\begin{equation}
\begin{aligned}
(g,\xi_1,\xi_2)\oplus(g',\xi_1',\xi_2'):=&(gg',\Ad_g(\xi_1')+\xi_1 \\
&-\ad_{\xi_2}(\Ad_g(\xi_2')),\Ad_g(\xi_2')+\xi_2)
\end{aligned}
	\label{}
\end{equation}
The identity element of this Lie group is $(I,0,0)$, and the inverse of $(g,\xi_1,\xi_2)\in\SE\times\se^2$ is given by:
\begin{equation}
	(g,\xi_1,\xi_2)^{-1}=(g^{-1},-\Ad_{g^{-1}}(\xi_1),-\Ad_{g^{-1}}(\xi_2))
	\label{}
\end{equation}
We may take advantage of this isomorphism by first scanning the isomorphic operands $(f_{i-1,i}^{-1},S_i\ddot q_i,S_i\dot q_i)$'s and then performing a parallel isomorphism to recover $V_i$'s and $\dot V_i$'s. 

\subsection{Backward scan of bias force and joint torques for inverse dynamics}

After scanning the link velocities and accelerations, the bias term $\hat F_i:=J_i\dot V_i-\ad_{V_i}^T(J_iV_i)$ in the first equation of \eqref{eq:id-backward} may be pre-computed in parallel before a second scan computes the linear recursion for $F_i$'s. Alternatively, $\hat F_i$ is quadratic in $V_i$ and linear in $\dot V_i$, and may be synchronously computed along with the scan of \eqref{eq:scan-id-vel-acc}.
Moreover, it can be proved that $\hat F_i$ involves only $9$ out of the $21$ quadratic terms of $V_i=(v_i^T,w_i^T)^T$, namely $w_{ij}w_{ik}$ for $1\leq j\leq k\leq 3$ and $w_i=(w_{i1},w_{i2},w_{i3})^T$, and $w_i\times v_i\in\mathds R^3$. For convenience, we define $Q_i=((w_i\times v_i)^T,w_{i1}^2,w_{i1}w_{i2},w_{i1}w_{i3},w_{i2}^2,w_{i2}w_{i3},w_{i3}^2)^T\in\mathds R^9$. A synchronous scan may be expressed as follows:
\begin{equation}
\begin{aligned}
	\begin{bmatrix}
		\dot V_i\\
		\hline
		Q_i\\
		V_i\\
		\hline
		\hat F_i\\
		\hline
		1
	\end{bmatrix}&=\underbrace{
		\left[
		\begin{array}{c|cc|c|c}
			*&0&*&0&*\\
			\hline
			0&*&*&0&*\\
			0&0&*&0&*\\
			\hline
			*&*&*&0&*\\
			\hline
			0&0&0&0&1 	
		\end{array}
	\right]}_{\displaystyle A_i\in\mathds R^{28\times 28}}
		\begin{bmatrix}
			\dot V_{i-1}\\
			\hline
			Q_{i-1}\\
			V_{i-1}\\
			\hline
			\hat F_{i-1}\\
			\hline
			1
	\end{bmatrix}\\
    F_0&=0
    \end{aligned}
	\label{eq:id-fwd-scan}
\end{equation}
where only the block pattern of the scan operand $A_i$ is shown. A similar simplification of the operand by isomorphism may be further investigated.
After scanning the forward propagation \eqref{eq:id-fwd-scan}, we proceed with the scan of the backward propagation \eqref{eq:id-backward} using the following linear recursion:
\begin{equation}
\begin{aligned}
	\begin{bmatrix}
		F_i\\
		\hline
		\tau_{i+1}\\
		\hline
		1
	\end{bmatrix}&=
	\underbrace{\left[
	\begin{array}{c|c|c}
		\Ad_{f_{i,i+1}^{-1}}^T & 0& \hat F_i\\
		\hline
		S_{i+1}^T & 0& 0 \\
		\hline
		0 & 0 & 1
	\end{array}
\right]}_{\displaystyle A_i\in\mathds R^{8\times 8}}
	\begin{bmatrix}
		F_{i+1}\\
		\hline
		\tau_{i+2}\\
		\hline
		1
	\end{bmatrix}\\
    F_0&=\tau_{n+1}=\tau_{n+2}=0
	\end{aligned}
	\label{eq:id-bkwd-scan}
\end{equation}
Therefore, we have shown that the ID problem may be completely solved by two scanning operations.

\vskip-10pt
\subsection{Accelerating JSIIA using scan operations}

Following the discussion about \eqref{eq:JSIinv} in Section \ref{sec:racap:FD}, we may accelerate the JSIIA by parallelizing data-independent computation of the columns $M_{\cdot,j}(q)$ of the JSI $M(q)$ (see~\cite[Ch. 6]{featherstone2014rigid} for more details) using our parallel ID algorithm:
\begin{equation}
	\begin{aligned}
	M_{\cdot,j}(q)=&M(q)\delta_{\cdot,j}\\
	=&\ID(q,\dot q,\delta_{\cdot,j},V_0,\dot V_0,F_{n+1})\\
    &-\ID(q,\dot q,0,V_0,\dot V_0,F_{n+1})\\
	=&\ID(q,0,\delta_{\cdot,j},0,0,0)\\
	\delta_{\cdot,j}:=&(\delta_{1,j},\cdots,\delta_{n,j})^T\quad j=1,\cdots,n
	\end{aligned}
	\label{}
\end{equation}
where $\delta_{i,j}$ denotes the Kronecker delta function. The joint accelerations $\ddot q$ may then be evaluated from \eqref{eq:JSIinv} using state-of-the-art parallel linear system solvers, such as parallel Cholesky decomposition~\cite{volkov2008lu}. Consequently, a total of $n+1$ parallel IDs (including the one for computing $\tau^{\text{bias}}$) are computed in parallel for the JSIIA.

\subsection{Accelerating ABIA using scan operations}

The details of square factorization of JSI and its analytical inversion may be found in \cite{ploen1999coordinate}. Following our discussion in Section \ref{sec:racap:FD}, the two-phase propagation summarized in \eqref{eq:recursive-fd} may be reformulated into a backforward scan:
\begin{equation}
\begin{aligned}
	\begin{bmatrix}
		\hat z_i\\
		\hat c_{i+1}\\
		\hline
		1
	\end{bmatrix}&=\underbrace{\left[
	\begin{array}{c|c|c}
		Y_{i,i+1} & 0 & \mathrm\Pi_{i,i+1}\hat\tau_{i+1}\\
		-\mathrm\Omega_{i+1}^{-1}S_{i+1}^T&0&\mathrm\Omega_{i+1}\hat\tau_{i+1}\\
		\hline
		0&0&1
	\end{array}
\right]}_{\displaystyle A_i\in\mathds R^{8\times 8}}	\begin{bmatrix}
		\hat z_{i+1}\\
		\hat c_{i+2}\\
		\hline
		1
	\end{bmatrix}\\
    \hat{z}_{n+1}&=\hat{c}_{n+1}=\hat{c}_{n+2}=0
    \end{aligned}
	\label{eq:fd-bkwd-scan}
\end{equation}
followed by a forward scan:
\begin{equation}
	\begin{aligned}
	\begin{bmatrix}
		\lambda_i\\
		\ddot q_i\\
		\hline
		1
	\end{bmatrix}&=\underbrace{\left[
	\begin{array}{c|c|c}
		Y_{i-1,i}^T&0&S_i\hat c_i\\
		-\mathrm\Pi_{i-1,i}^T&0&\hat c_i\\
		\hline
		0&0&1
	\end{array}
	\right]}_{\displaystyle A_i\in\mathds R^{8\times 8}}	\begin{bmatrix}
		\lambda_{i-1}\\
		\ddot q_{i-1}\\
		\hline
		1
	\end{bmatrix}\\
    \lambda_0&=\ddot q_0=0
	\end{aligned}    
	\label{eq:fd-fwd-scan}
\end{equation}
respectively. Therefore, the complete ABIA comprises two scan operations with operands $A_i$'s given in \eqref{eq:id-fwd-scan},\eqref{eq:id-bkwd-scan} for the computation of bias torques $\tau_i^{\text{bias}}$'s, a nonlinear recursion \eqref{eq:abi} for the computation of the ABI, and then finally two scan operations with operands $A_i$'s given in \eqref{eq:fd-bkwd-scan},\eqref{eq:fd-fwd-scan} for the computation of joint accelerations $\ddot q_i$'s. We may also combine the two backward scans pertaining to \eqref{eq:id-bkwd-scan} and \eqref{eq:fd-bkwd-scan} into a single backward scan, with the corresponding linear recursion given by:
\begin{equation}
\begin{aligned}
	\begin{bmatrix}
		F_{i-1}\\
		\hat\tau_i\\
		\hline
		\hat z_i\\
		\hat c_{i+1}\\
		\hline
		1
	\end{bmatrix}=&\underbrace{\left[
	\begin{array}{c|cc|c|c}
		\Ad_{f_{i-1,i}^{-1}}^T & 0 & 0 & 0 & \hat F_{i-1}\\
		-S_i^T & 0 & 0 & 0 & \tau_i^{\text{in}}\\
		\hline
		0& \mathrm\Pi_{i,i+1} & Y_{i,i+1} & 0 & 0\\
		0&\mathrm\Omega_{i+1}& -\mathrm\Omega_{i+1}^{-1}S_{i+1}^T&0&0\\
		\hline
		0&0&0&0&1
	\end{array}
\right]}_{\displaystyle A_i\in\mathds R^{15\times 15}} 
\\
 & \times \begin{bmatrix}
		F_i\\
		\hat\tau_{i+1}\\
		\hline
		\hat z_{i+1}\\
		\hat c_{i+2}\\
		\hline
		1
	\end{bmatrix}
 \end{aligned}
	\label{eq:fd-bkwd-scan-large}
\end{equation}
thereby reducing the hybrid ABIA to one nonlinear recursion to be executed on the CPU, and a sequence of three parallel scans to be executed on the GPU.
\section{Implementation}
\label{sec:gpu-impl}

So far, we have presented a complete parallel framework for utilizing parallel scans to accelerate articulated robot ID and FD computation. In comparison to the classic recursive algorithms \cite{ploen1999coordinate}, we have eliminated a large portion of intermediate variables by assembling multiple linear recursions to a minimal number of large dimension linear recursions (two for ID, and three for ABIA). However, such mathematically compact form does not necessarily lead to an optimal data structure for implementation on a particular hardware platform. In this section, we shall materialize the corresponding parallel algorithms on a hybrid CPU-GPU platform, and illustrate how hardware constraints may weigh in to tailor our algorithms with the flexibility of composing/decomposing linear recursions. We adopt Nvidia CUDA as our software to make our implementation may be easily reproducible for further study and comparison.

\subsection{Parallel Inverse Dynamics Algorithm}

In Section~\ref{sec:recap:ID}, we proposed a forward-backward double scan parallelization of the recursive ID algorithm~\cite{park1995lie}, where the forward scan computes all link velocities and accelerations (along with the bias force) and the backward scan other for computing bias forces. In the actual GPU implementation, we split the scan for velocities and accelerations into two successive scans so that the scan data may by fitted into the GPU's highspeed shared memory and local registers. The computation of bias forces $\hat F_i$'s becomes perfectly parallel after the velocity and acceleration scans, may simply be computed on $n$ GPU threads, with $n$ being the number of links. The details of our parallel ID algorithm are illustrated in Algorithm~\ref{alg:par_id_alg}. 

\begin{algorithm}[!h]
	\caption{Parallel Inverse Dynamics $\textsc{CalcInvDyn}([M_i], [q_i],[\dot{q}_i],[\ddot{q}_i])$}
	\label{alg:par_id_alg}
	\begin{algorithmic}[1]
		\PARCMPT Compute adjacent transformations in parallel.
		\STATE $[f_{i-1,i}] \gets \textsc{CalcTransform}([M_i],[S_i],[q_i]) $
		\FWSCAN Compute body velocities.
		\STATE $ [V_i] \gets \textsc{InclusiveVelScan}([f_{i-1,i}],[S_i],[\dot{q}_i]) $
		\FWSCAN Compute body accelerations.
		\STATE $ [\dot{V}_i] \gets \textsc{InclusiveAccScan}([f_{i-1,i}],[S_i], [\ddot{q}_i],[V_i]) $
		\BWSCAN Compute body forces.
		\STATE $ [F_i] \gets \textsc{InclusiveForceScan}([f_{i-1,i}], [V_i], [\dot{V_i}]) $
		\PARCMPT Compute joint torques.
		\STATE $ [\tau_i] \gets \textsc{CalcTorque}([S_i],[F_i]) $
	\end{algorithmic}
\end{algorithm}

\subsection{Forward Dynamics Algorithm}

As shown in Section~\ref{sec:dyn-scan}, the parallelization of FD algorithms such as JSIIA and ABIA relies on
recognizing the parallel ID algorithm as a common subroutine to compute the bias torques and/or the inertia matrix.
Although, neither FD algorithms are completely scannable, the JSIIA may be alternatively accelerated by parallelly solving a positive definite linear system of equations. The ABIA, on the other hand, relies on CPU recursion to compute the ABI, and may be accelerated by careful scheduling of the hybrid CPU/GPU system.

\subsubsection{Parallel JSIIA}

Given a robot with $n$ links, JSIIA calls the inverse dynamics routine $n+1$ times with different inputs. The first call computes the bias torques $\tau^{\text{bias}}_i$'s by setting zero-acceleration input, i.e., $\ddot{q}_i = 0, i=1,\dots,n$. 
The other $n$ inverse dynamics calls compute the $n$ columns of the JSI $M(q)$  by setting $\dot{q}=0$ and $\ddot{q} = \delta_{\cdot,j}, j=1,\dots,n$. All $n+1$ inverse dynamics are data independent, and thus may be solved simultaneously on GPUs.
Besides, the additional matrix operations involved in the JSIIA can also be computed in parallel on GPUs for further acceleration. The details of our parallel JSIIA are illustrated in Algorithm~\ref{alg:JSI}. 

\begin{algorithm}[!h]
	\caption{Parallel JSIIA $\textsc{CalcFwdDyn\_JSIIA}([\tau^{\text{in}}_i], [q_i], [\dot{q}_i])$}
	\begin{algorithmic}[1]
		\PARCMPT Compute the bias torques and the joint space inertia matrix.
		\STATE $\begin{aligned} ([\tau^{\text{bias}}_i],\ M_{\cdot,1,...,n}) \gets
         (&\textsc{CalcInvDyn}([q_i], [\dot{q}_i], [0]), \\
         &\textsc{CalcInvDyn}([q_i], [0], [\delta_{i,1}]), \\
         &..., \\
         &\textsc{CalcInvDyn}([q_i], [0], [\delta_{i,n}])
         \end{aligned}$ 
		\PARCMPT Compute differential torques.
		\STATE $ [\tau^{\text{diff}}_i] \gets \textsc{CalcDiffTorques}([\tau^{\text{in}}_i],\ [\tau^{\text{bias}}_i]) $
		\PARCMPT Compute the inverse of the joint space inertia matrix.
		\STATE $M^{-1} \gets \textsc{CalcMatrixInv}(M) $
		\PARCMPT Compute joint accelerations.
		\STATE $ [\ddot{q}^{\text{out}}_i] \gets \textsc{CalcAcc}(M^{-1},\ [\tau^{\text{diff}}_i]) $
	\end{algorithmic}
	\label{alg:JSI}
\end{algorithm}

\subsubsection{Hybrid ABIA Algorithm}

In comparison to JSIIA, ABIA only leverages the inverse dynamics routine once for bias torque computation. The ABI $\hat J_i$'s is recursively computed using~\eqref{eq:abi} on the CPU and sent back to the GPUs for subsequent forward dynamics computation. The high latency of sending data from CPU main memory to the GPU video memory is partially hidden through asynchronous computation of ABIs and bias torques. All remaining components can be performed in parallel on GPUs. This includes the computation for several intermediate variables, which are either perfectly parallel (line 4, 6, 8 in Algorithm~\ref{alg:hybrid_abi}) or scannable (line 5, 7 in Algorithm~\ref{alg:hybrid_abi}). The complete description of the hybrid ABIA algorithm is shown in Algorithm~\ref{alg:hybrid_abi}.

\begin{algorithm}
	\caption{Hybrid ABIA Algorithm $\textsc{CalcFwdDyn\_ABIA}([\tau_{\text{in}_i}], [q_i], [\dot{q}_i])$}
	\label{alg:hybrid_abi}
	\begin{algorithmic}[1]
		\HETER
		\STATE Compute bias torques.
        \begin{align*}
        	[\tau^{\text{bias}}_i] \gets \textsc{CalcInvDyn}([q_i], [\dot{q}_i], [0])
		\end{align*}
		\STATE Compute differential torques.
		\begin{align*}
        [\tau^{\text{diff}}_i] \gets \textsc{CalcDiffTorques}([\tau^{\text{in}}_i],[\tau^{\text{bias}}_i])
		\end{align*}
		\STATE Compute articulated body inertia on the CPU.
		\begin{align*}
        [\hat{J}_i] \gets \textsc{CalcABI}([J_i], [f_{i-1,i}], [S_i])
		\end{align*}
		\STREAM Parallel compute intermediate variables on the GPU. \\
        \STATE $ \begin{aligned} 
        &\left( [\mathrm\Omega_i], [\mathrm\Pi_{i, i+1}], [Y_{i,i+1}]\right) \\
        & \ \ \ \gets \textsc{CalcInterIntVar}([\hat{J}_i], [f_{i-1,i}], [S_i]) 
        \end{aligned}$
		\BWSCAN Compute intermediate variables $ [\hat{z}_i] $.
		\STATE $ [\hat{z}_i] \gets \textsc{InclusiveZhatScan}([Y_{i,i+1}], [\mathrm\Pi_{i, i+1}], [\tau^{\text{diff}}_i]) $ 
		\PARCMPT Compute intermediate variables $ [\hat{c}_i] $.
		\STATE $ [\hat{c}_i] \gets \textsc{CalcChat}([\hat{z}_i],\ [S_i],\ [\tau^{\text{diff}}_i]) $
		\FWSCAN Compute intermediate variables $ [(\lambda_i)] $.
		\STATE $ [\lambda_i] \gets \textsc{InclusiveLambdaScan}([Y_{i,i+1}],\ [S_i],\ [\hat{c}_i]) $
		\PARCMPT Compute joint accelerations.
		\STATE $ [\ddot{q}^{\text{out}}_i] \gets \textsc{CalcAcc}([\lambda_i],\ [\hat{c}_i],\ [\mathrm\Pi_{i, i+1}]) $
	\end{algorithmic}
\end{algorithm}

\section{Experiments}
\label{sec:experiment}

In this section, we evaluate the performance of our GPU-based parallel dynamics algorithms by comparing their running time to that of their multithreading CPU counterparts on two experiments. The first experiment investigates the speedup and scalability with respect to the number of links of a single robot. 
The second one is used to demonstrate the efficacy of our methods on a large group of  robots with moderate number of links, which poses as a bottleneck in many applications such as model-based control optimization. The robot configuration parameters such as twists are initialized randomly before the dynamics computation in all our experiments. 

Each experiment will be repeated 1000 times with randomized joint inputs, and the average computation time is then reported. All running times are recorded on a desktop workstation with an 8-core Genuine Intel i7-6700 CPU and 15.6 GB memory. We implemented our GPU-based parallel dynamics algorithms using CUDA on a Tesla K40c GPU with a 11520 MB video memory and 288GB/sec memory bandwidth. CPU multithreading is used to parallelize the large number of independent dynamics computations in the second set of experiments, where the number of threads is equal to the number of independent dynamics computations.

\subsection{Experiments with Different Link Numbers}

\begin{figure}
	\subfigure[Comparison of ID]{\label{fig:comparison_id}
    \includegraphics[width=.5\textwidth]{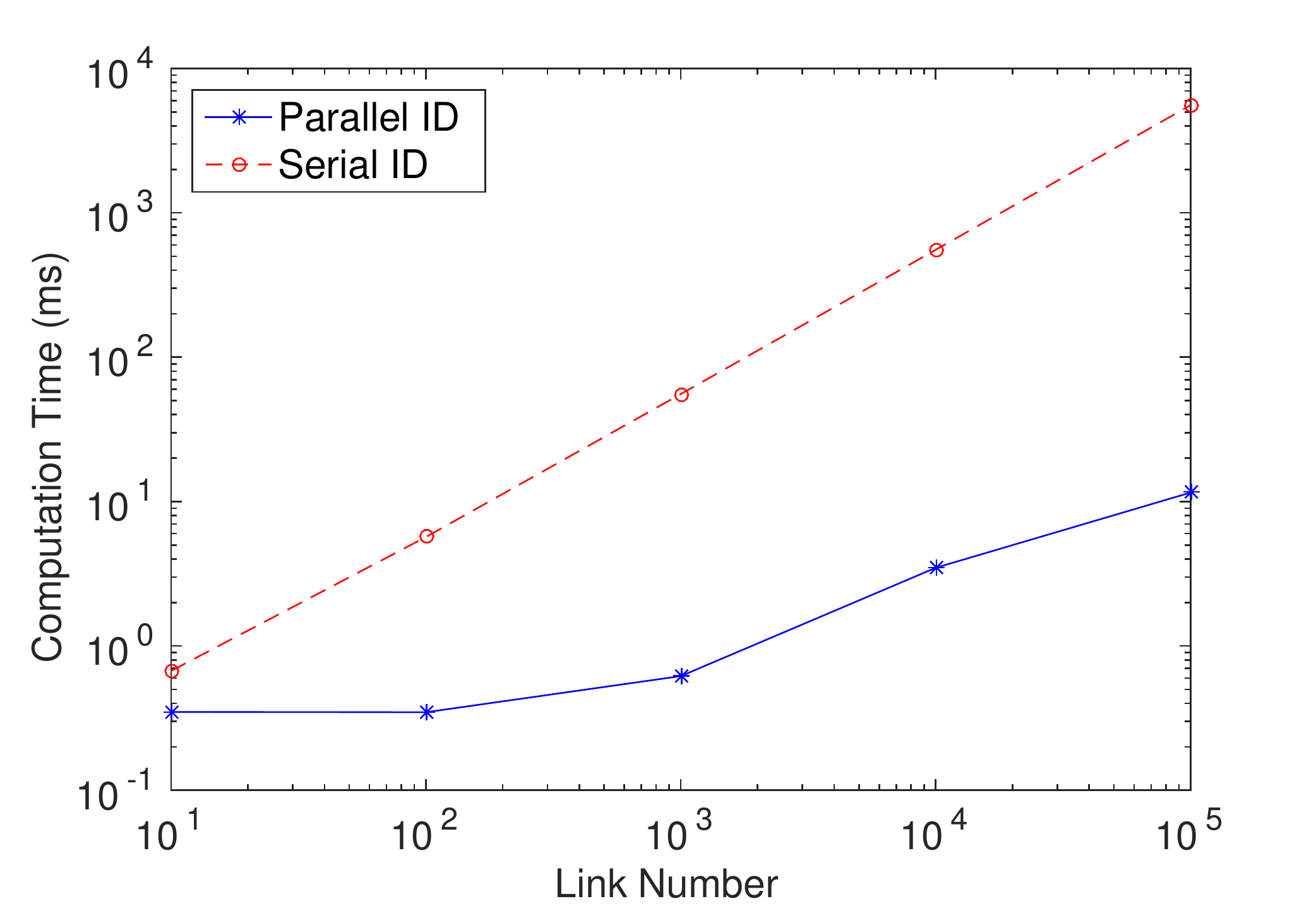}}
    \subfigure[Comparison of FD]{\label{fig:comparison_jsi_abi}
    \includegraphics[width=.5\textwidth]{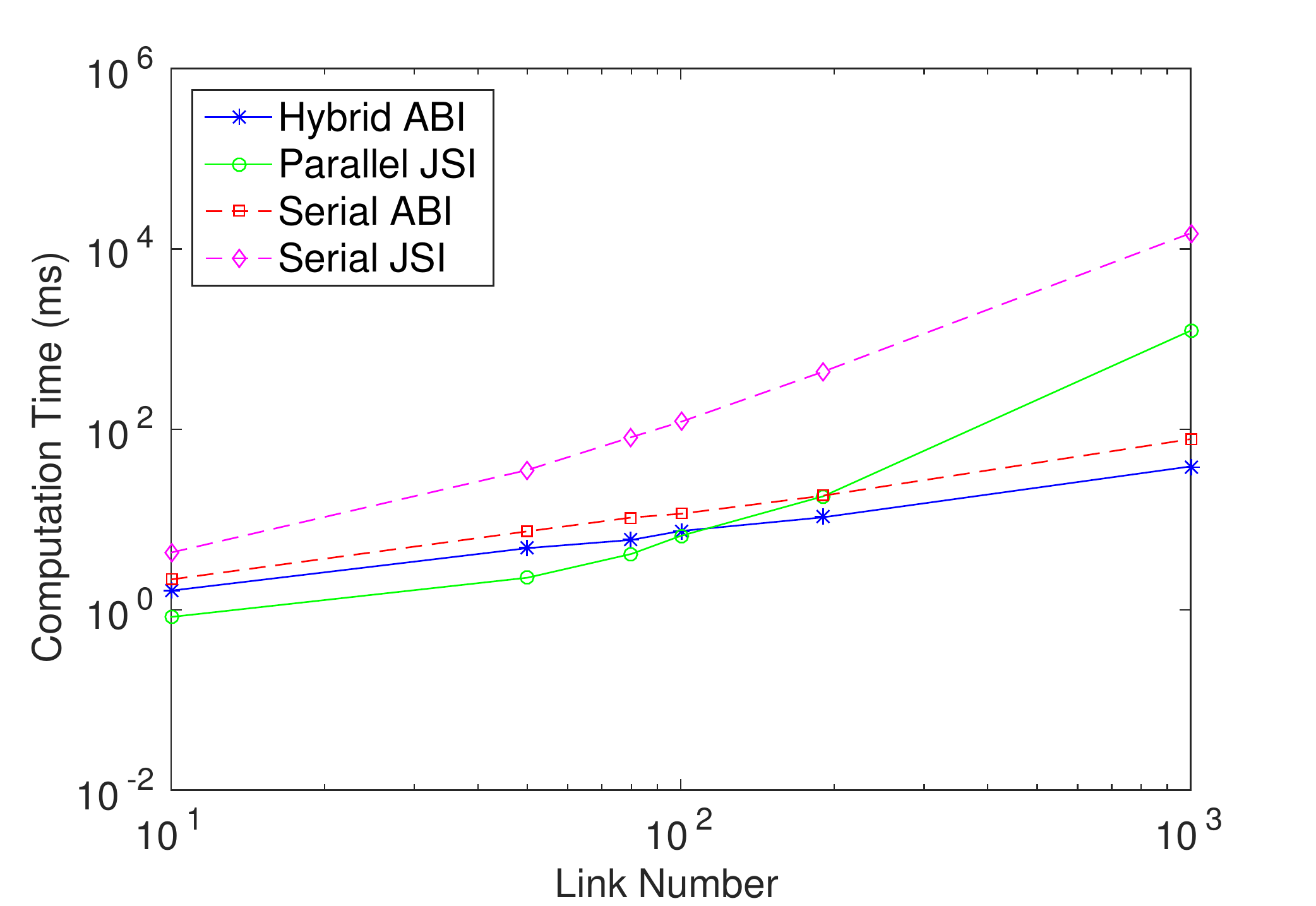}}
  \caption{Computation time comparison of different algorithms on a single inverse / forward dynamics call for robots with different number of links}
  \label{fig: link_diff}  
\end{figure}

We first compare the GPU and CPU's performance of computing a single inverse dynamics for robots with different number of links, and the result is shown in Figure~\ref{fig:comparison_id}.
We can observe that when the number of links increases, the serial CPU's computation time increases linearly, while the GPU's time cost roughly increases at the $\log (n)$ rate.

Next, we compare the time cost of a single forward dynamics call for robots with different number of links. We investigate four different implementations of the forward dynamics computation, including GPU parallel JSIIA, CPU-GPU hybrid ABIA, serial JSIIA and serial ABIA. According to the result shown in Figure~\ref{fig:comparison_jsi_abi}, the serial JSIIA has the worst performance and is dominated by other three methods. The GPU-based parallel JSIIA is the fastest for robots with moderate number of links, but it is outperformed by hybrid ABIA and the serial ABIA when the link the number is more than $100$ links or $190$ links respectively. This is because the JSIIA involves the
inversion of inertia matrix. For robots with a large number of links, the inertia matrix inversion is more computationally expensive than any components in the ABIA algorithm, and may involve expensive data exchange between global memory and local memory. Our hybrid ABIA algorithm will be the most efficient for robots with many links, thanks to the asynchronous data exchange which hides the latency of sending data from CPUs to GPUs.

\subsection{Experiments with Different Group Numbers}
We now compare the performance while performing a large number of independent dynamics computations between GPU-based parallel algorithms and multithreading CPU algorithms. In the CPU implementation, we allocate one thread for each independent dynamics computation. 

\begin{figure}
  \subfigure[Comparison for robots with 10 links]{\label{fig:id_10}
  \includegraphics[width=.5\textwidth]{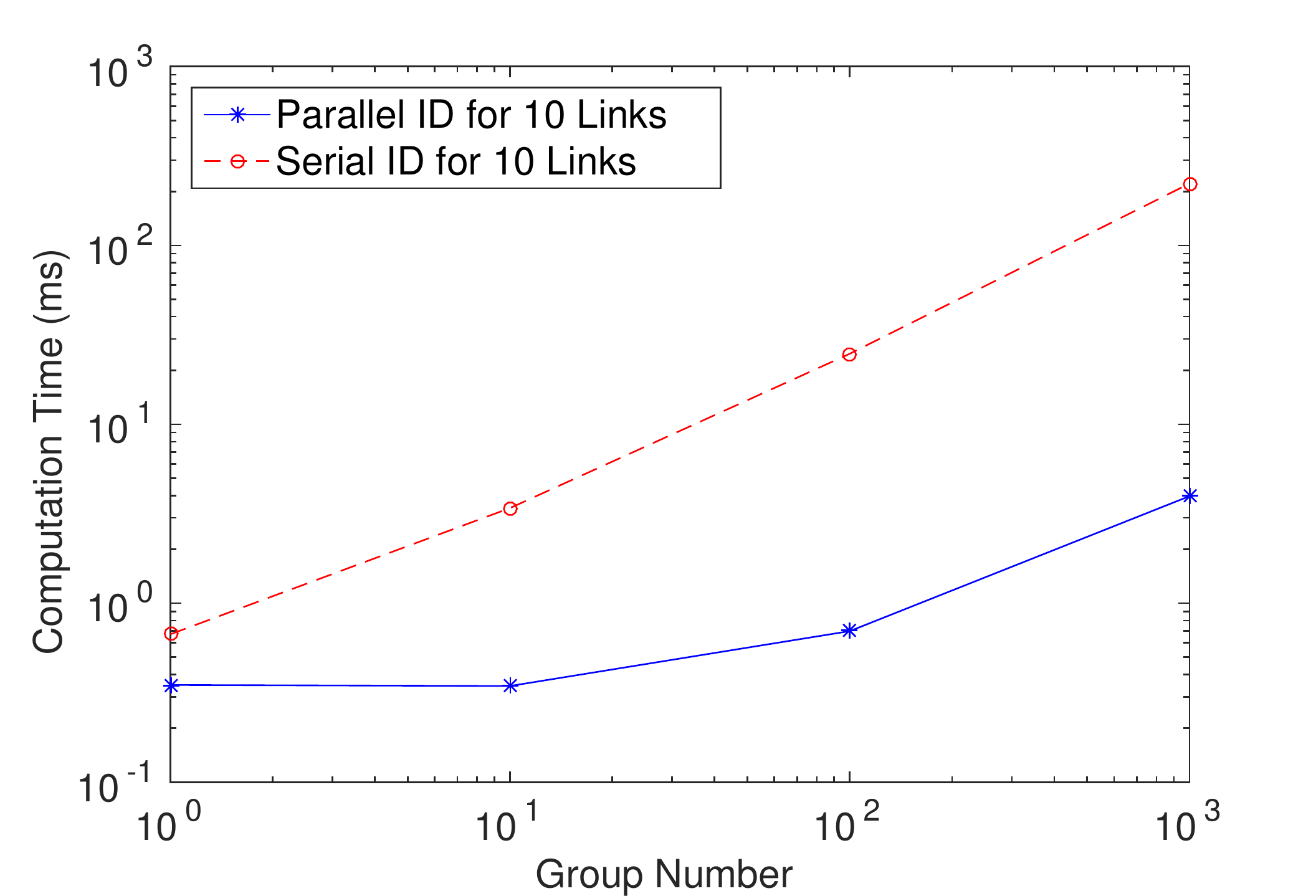}}
  \subfigure[Comparison for robots with 100 links]{\label{fig:id_100}
  \includegraphics[width=.5\textwidth]{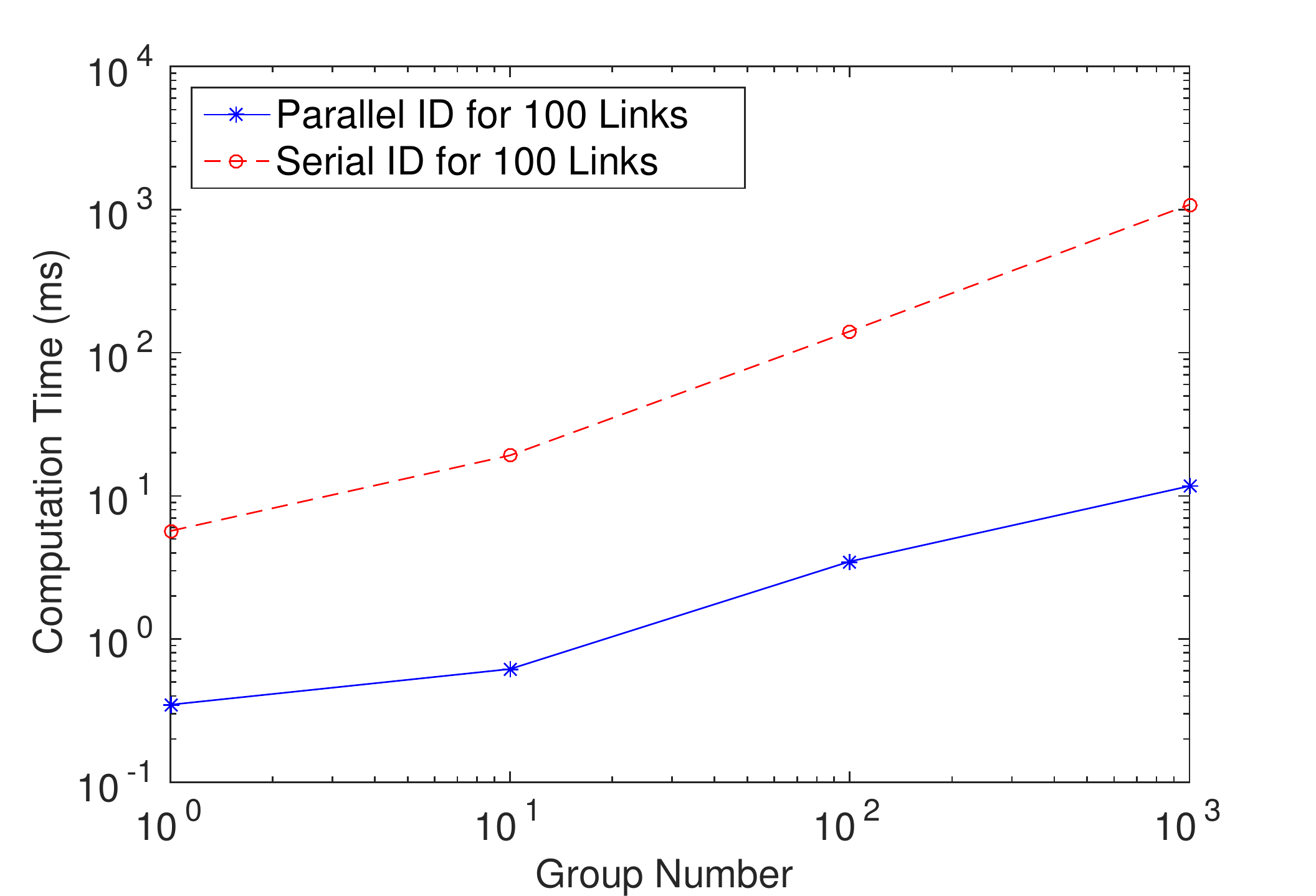}}
  \caption{Computation time comparison of different algorithms on many independent inverse dynamics call}
  \label{fig: id_grp}
\end{figure}

The comparison result for inverse dynamics implementations are shown in Figure~\ref{fig:id_10} and Figure~\ref{fig:id_100} for robots with 10 links and 100 links respectively. We see that the running time of the CPU inverse dynamics increases linearly with the group number. In contrast, the computation time of the GPU inverse dynamics increases much slower. Due to the limited memory and necessary thread management of the GPU, the GPU inverse dynamics does not reach the ideal $O(\log(n)) $ time complexity. Nevertheless, it provides a satisfactory $\sim$100x speedup when the group number is 1000.

\begin{figure}
  \subfigure[Comparison for robots with 10 links]{\label{fig:fd_10}
    \includegraphics[width=.5\textwidth]{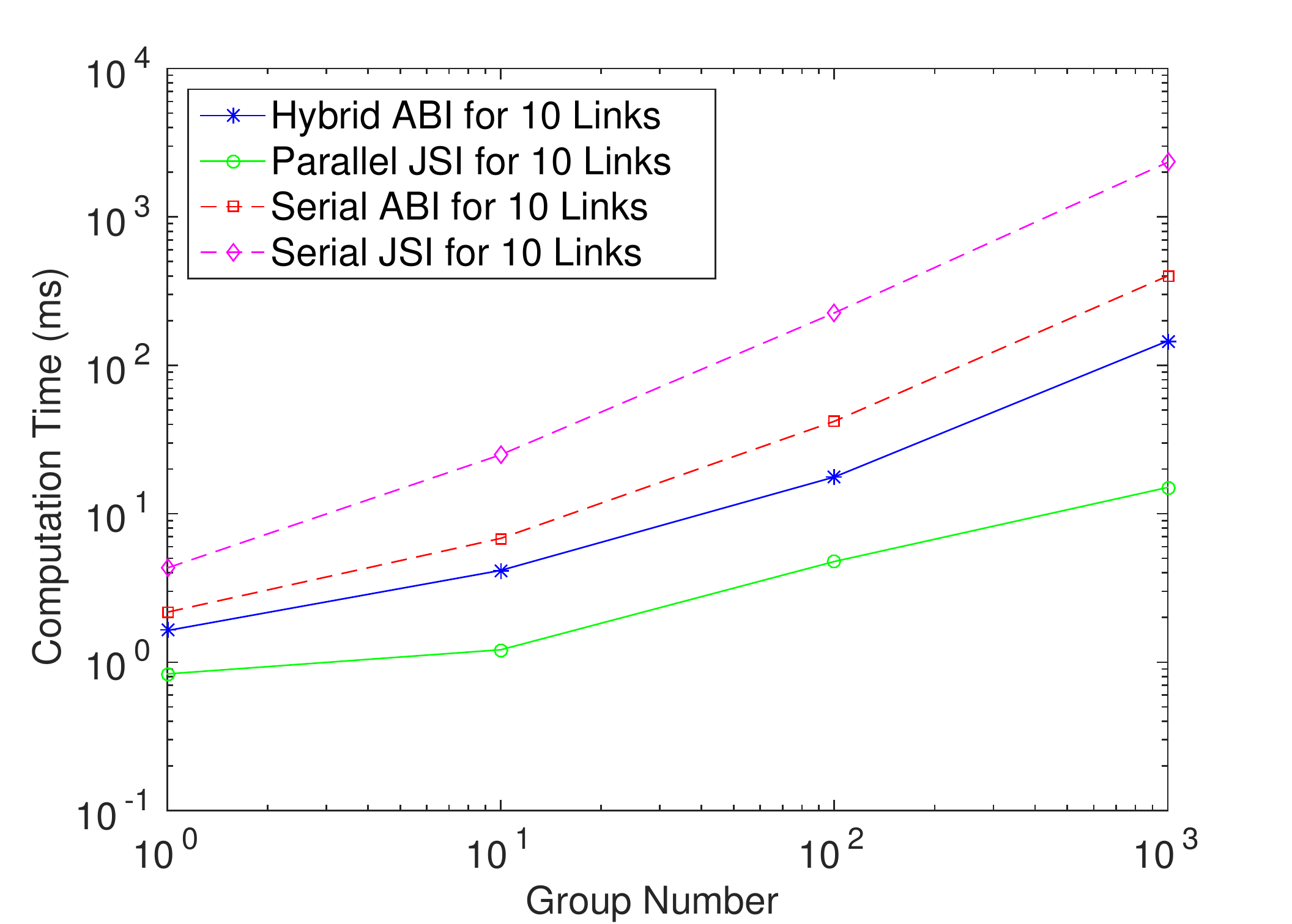}}
  \subfigure[Comparison for robots with 200 links]{\label{fig:fd_200}
    \includegraphics[width=.5\textwidth]{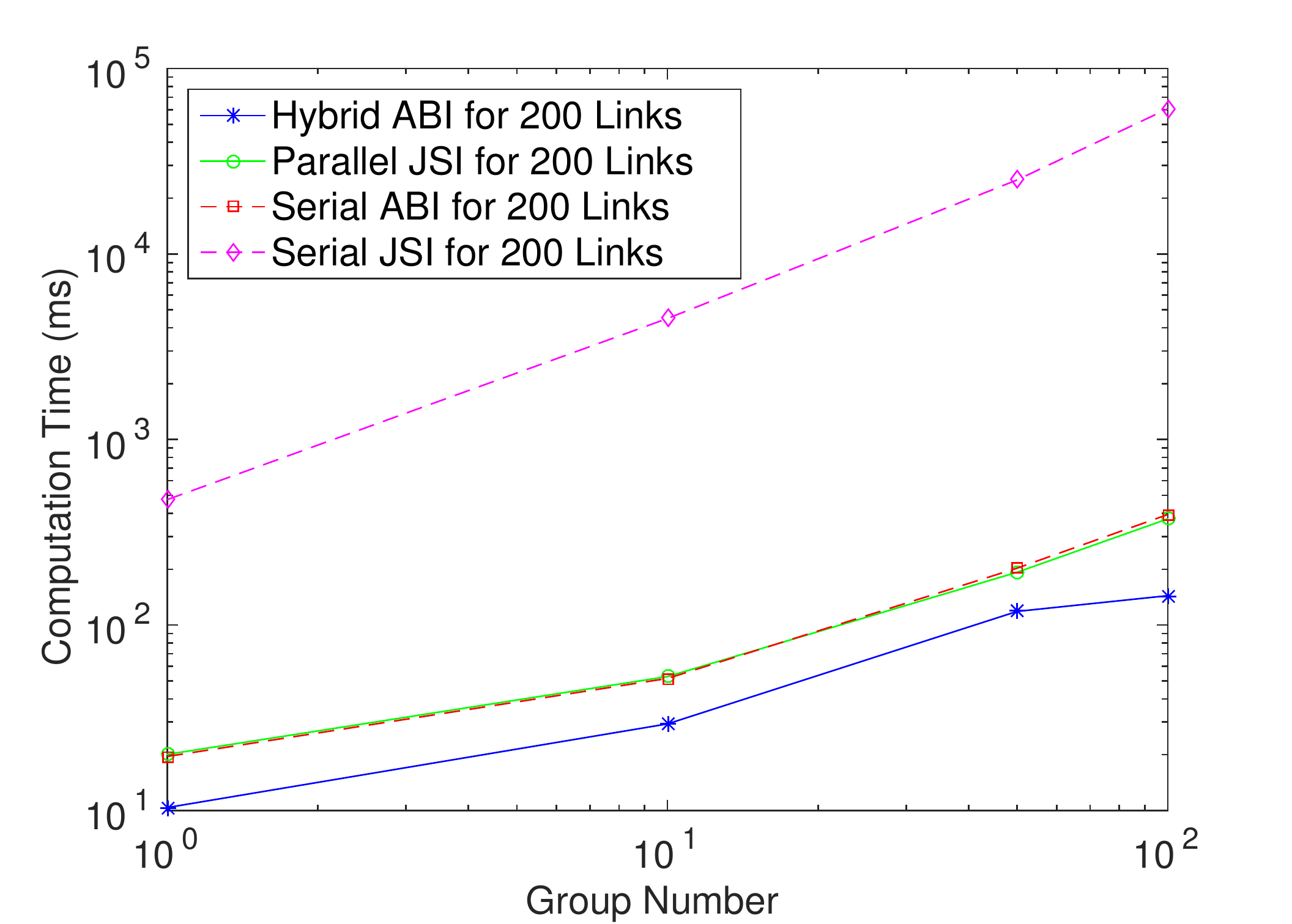}}
  \caption{Computation time comparison of different algorithms on many independent forward dynamics call}
  \label{fig:fd_grp}
\end{figure}

Next, four different approaches for computing the forward dynamics are compared, namely GPU-based parallel JSIIA and ABIA, CPU multithreading JSIIA and ABIA, and the results are shown in Figure~\ref{fig:fd_grp}. From Figure~\ref{fig:fd_10}, we can observe that the GPU-based JSIIA is always the most efficient one. When the link number is $200$, the result is quite different as shown in Figure~\ref{fig:fd_200}: the hybrid ABIA always outperforms the other three methods, and the multithreading JSIIA is the slowest approach. This phenomenon is consistent with the result in Figure~\ref{fig:comparison_jsi_abi} where the GPU-based JSIIA and the hybrid ABIA have the shortest running time when the link number is 10 and 200 respectively. The reason is because when the link number is small (e.g., 10), the inertia matrix is small enough to be fit in the high-speed memory of the GPUs for high performance; and we can further accelerate the dynamics computation by parallelizing the matrix inversions in independent dynamics calls. When the link number is large (e.g., 200), the inertia matrix becomes so large that the running time for a single matrix inversion is significantly more expensive than other components of the dynamics computation. Even worse, due to the limited GPU memories, we have to perform these time-consuming matrix inversions in serial and can not leverage GPUs' parallel mechanism. 

\section{Conclusion and Future Work}

In this paper, we presented a parallel framework for fast computation of inverse and forward dynamics of articulated robots using parallel scan operations, and reported details of its implementation on a hybrid CPU-GPU platform. Our contributions are summarized as follows:

\begin{enumerate}
	\item We reformulated the recursive Newton-Euler inverse dynamics algorithm into a forward-backward two-phase parallel scan operation, which is based on the utilization of several generalizations of the standard scan operation.
    \item We reformulated the joint space inertia inversion algorithm (JSIIA) into $n+1$ data-independent scan operations along with a positive definite matrix inversion operation.
    \item We reformulated the complete articulated-body inertia algorithm (ABIA) into a nonlinear recursion and a forward-backward-forward three-phase parallel scan operation.
    \item We implemented the aforesaid algorithms on a hybrid CPU-GPU platform with extensive use of Nvidia CUDA standard libraries. This makes our experiment results easily reproducible for further study and comparison.
\end{enumerate}
Besides, since our parallel framework is based on a coordinate-free Lie group / semigroup formulation and is not representation-specific, the corresponding parallel algorithms apply equally well to other representations than matrices, such as dual quaternions. 

Our future work shall involve expansion of our parallel framework to include parallel scan of gradient and Hessian of inverse dynamics \cite{lee2005newton}, and other parallel FD algorithms such as CFA \cite{fijany1995parallel}, and eventually an automatic scheduler for hardware-specific optimal performance.

\section*{Acknowledgments}

This work is partially supported by NVidia Corp. and Hong Kong GRF 17204115. Yuanqing Wu is supported by the PRIN 2012 grant No. 20124SMZ88 and the MANET FP7-PEOPLE-ITN grant No. 607643.

{\small
\bibliographystyle{IEEEtran}
\bibliography{reference}
}

\end{document}